\documentclass{article}

\usepackage[english]{babel}
\usepackage[section]{placeins}

\usepackage[letterpaper,top=2cm,bottom=2cm,left=3cm,right=3cm,marginparwidth=1.75cm]{geometry}

\usepackage{amsmath}
\usepackage{graphicx}
\usepackage[colorlinks=true, allcolors=blue]{hyperref}

\title{SARS-CoV-2 Result Interpretation based on Image Analysis of Lateral Flow Devices}
\author{Neeraj Vashistha}
\date{}
\begin{document}
\maketitle

\begin{abstract}
The widely used gene quantisation technique, Lateral Flow Device (LFD), is now commonly used to detect the presence of SARS-CoV-2. It is enabling the control and prevention of the spread of the virus. Depending on the viral load, LFD have different sensitivity and self-test for normal user present additional challenge to interpret the result. With the evolution of machine learning algorithms, image processing and analysis has seen unprecedented growth. In this interdisciplinary study, we employ novel image analysis methods of computer vision and machine learning field to study visual features of the control region of LFD. Here, we automatically derive results for any image containing LFD into positive, negative or inconclusive. This will reduce the burden of human involvement of health workers and perception bias. 

\end{abstract}

\section{Introduction}

The most common test to detect the presence of SARS-Cov2 are rapid antigen tests using LFD and real-time quantitative polymerase chain reaction(RT-qPCR). However, rapid antigen tests are the most cost-effective measure after the RT-qPCR test. LFD using rapid antigen test provides users with a simple, binary result (positive, negative) in as little as 5 to 20 minutes. These tests have been widely proliferated since the beginning of SARS-Cov2 and are commonly used to test pregnancy, malaria, H5N1, SARS, antibiotic resistance and legionella \cite{Wang2016}  The LDF contain a strip that is coated with a specific antigen on which human fluid (blood, plasma, mucus or urine) along with a reagent is placed. This strip contains two bands, (control band and test band) which appear after 5 mins to 20 mins. \cite{Amerongen2018} 

Although, RT-qPCR is considered the more accurate but the cost associated with it considering human efforts and equipment is very high. Rapid antigen tests on the other hand can detect viruses (both asymptomatic and symptomatic) at par with RT-qPCR tests\cite{Lee2021}. Rapid antigen tests are simple to use, relatively cheap and require minimal training. Both type of tests have helped in the suppression and spread of viruses within the UK boroughs \cite{Martin2021}. Two studies in 2020, showed LFD results when compared with PCR test showed the specificity is almost 99.9\% with sensitivity in the test to be around 48.9\% to 78.9\%\cite{Peto2021} and LFD positivity is directly proportional with viral load\cite{Marta2021}.

A potential issue with the use of LFD is interpreting the test results\cite{Isabela2020}. As previously mentioned, if the count of antibodies is less, the band colours in LFD are faintly visible. This sometimes tempts users to falsely read a test as negative or sometimes the eye acuity reads the test result wrong.  This task of interpretation becomes more complex in very elderly or young individuals, which can waste a lot of effort of mass testing and virus suppression.

The first objective of this study is to find a simple and intuitive way to determine the presence of the control band and the test band in the LFD result image using state-of-art machine learning models. The second objective is to scale this system both vertically and horizontally, which would mean that it would work for multiple retail vendors which are providing Lateral Flow kits in the market as well as can sustain a multitude of verification -requests at the same time. Third, reduce the human dependence for some borderline results which are too small to detect by using active learning techniques.

To achieve the above laid out objectives, we propose a multi-model machine learning strategy. These machine learning models are built in three phases. In the first phase, the models are trained on different collections of images of the Lateral Flow test results, these images are curated from labs and different experiments to represent real-world data. In the second phase, we use the result models to test on actual real-world images and tune the parameter settings of each of these models. This enables the models to understand the distribution of data in a real-world scenario. In the third phase, we use these tested models and deploy them to infer in real-time to classify a Lateral Flow test image result into positive, negative or inconclusive.

\section{Model Design}

\begin{figure}[!htb]
\centering
\includegraphics[width=0.9\textwidth]{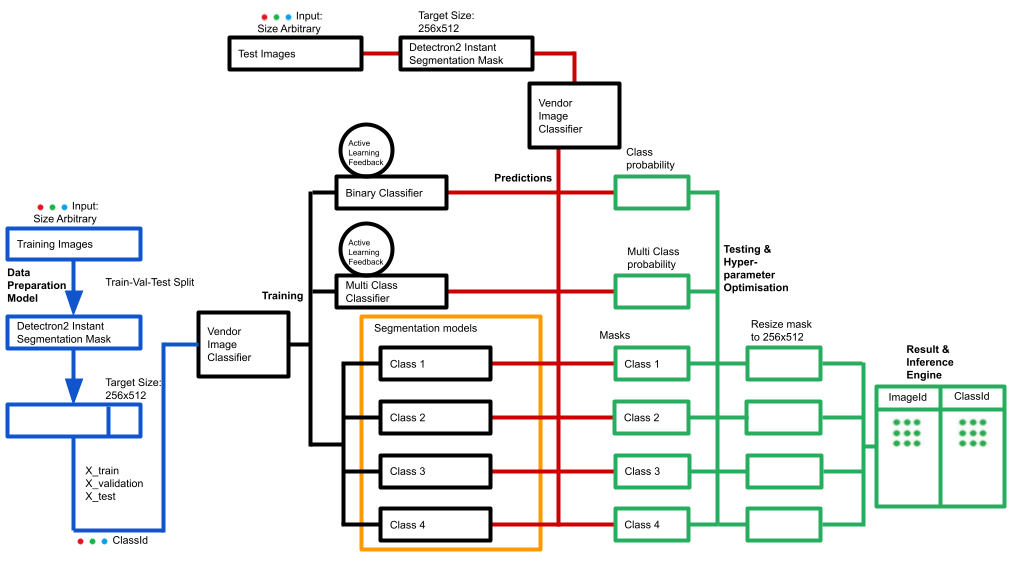}
\caption{\label{fig:model_arch} Machine Learning Model Architecture, the left section in blue represents the data preparation stage, the left-centre section in black represents Training and model development stage, the centre section in red represents testing phase, the right section in green represents Testing with validation, hyperparameter optimisation and inference stage.}
\end{figure}

The proposed multi-model machine learning framework is illustrated in Figure \ref{fig:model_arch}. Here, multiple arbitrary sized images containing Lateral Flow results were processed in the data preparation stage to extract the Region-of-Interest(ROI) part using Detectron2\cite{wu2019detectron2} instance segmentation based Mask R-CNN model\cite{he2018mask}. This results in a uniform fixed-size image extraction of the test strip which is only a selected portion of the test cassette. 

On these uniformed test strip image datasets, we train various classification models and segmentation models, such as vendor recognition image classifier, a binary and multi-class classifier with active learning component enabled. The binary classification models provide a probabilistic result for each test strip image as positive, negative or inconclusive. The multi-class classifier provides a more detailed probabilistic result for each test strip image giving results in terms of different classes such as Positive IGG, Positive IGM, Positive IGG and IGM, Negative or inconclusive, this multi-class classifier model can be tuned according to different vendors. 

The segmentation models\cite{Yakubovskiy2019} are a set of validation models for the above classifier models. In computer vision, image segmentation is the process of partitioning a digital image into multiple segments (sets of pixels, also known as image objects). The goal of segmentation is to simplify and/or change the representation of an image into something more meaningful and easier to analyze. Image segmentation is typically used to locate objects and boundaries (lines, curves, etc.) in images. To support the classification models, we use the result of the segmentation models and integrate it to provide a conclusive result. Since the segmentation model has to work in coordination with the classification model, the parameter tuning of segmentation model and loss function tuning is performed only after classification model parameters are tuned. 

After hyperparameter tuning of classification model and segmentation model is completed, these models are optimised for scalability which enables them to be run in parallel and inference at real time. Sometimes classification models are sufficient if the probabilistic scores for a class are above a certain threshold, in such case segmentation models are not inferred to save the memory and inference time. Thus, each model is configurable and independent of each other with certain constraints of the overall architect design. 

\section{Data}

Data used in this work is curated manually at Medusa19 and Avacta labs, it consists of only image samples and associated test results. It is the only information required for the development, training, testing and validation of machine learning models. Table 1, describe the overall data on which this study was performed.
\begin{table}[!htb]
\centering
\begin{tabular}{l|r|r}
 Type & Medusa19 & Avacta \\\hline
Positive images & 306, 77.2\% & 18, 10.4\% \\
Negative images & 104, 22.1\% & 153, 88.4\% \\
Inconclusive images & 3, 0.6\% & 2, 1.15\%\\
\end{tabular}
\caption{\label{tab:widgets}Training Samples for model building}
\end{table}
\begin{figure}[!htb]
\centering
\includegraphics[width=0.7\textwidth]{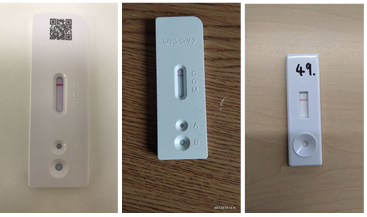}
\caption{\label{fig:test_strips} Training Sample Images, left Medusa19 with QR code, center Medusa19 w/o QR code and right Avacta w/o QR code.}
\end{figure}

The raw images can come in various sizes and shapes with different orientation but the default configuration are, portrait mode, with acceptable angle of rotation of the test cassette from 0 to +-18 degrees. Figure \ref{fig:test_strips}, shows some of the curated raw images.

\section{Model Development}

There are four models which effectively work in this system. The Instant Segmentation model, Classification Models, Image Segmentation Model and Active Learning model. The instant Segmentation model and classification model are one of the most critical models, although independent of each other but these need to work one after another, and loaded parallelly into the memory as the result of the first is consumed in the latter. Image Segmentation Model and Active Learning models improve the results of the other two models. In the below section, we will describe how each model is built, used and its performance metrics.

\subsection{Instant Segmentation Model}

\begin{figure}[!htb]
\centering
\includegraphics[width=0.6\textwidth]{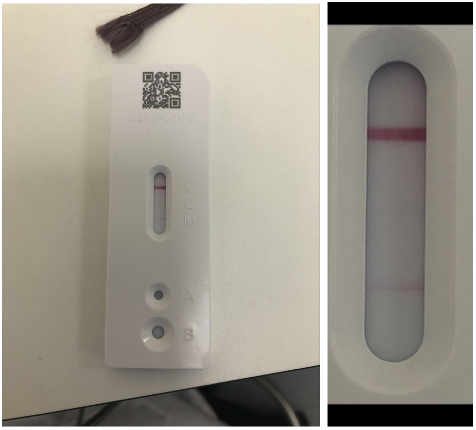}
\caption{\label{fig:figure3} Left raw Medusa19 test cassette image (2722 x 3607) and right test strip extract (300 x 875)}
\end{figure}

This model is built on an instance segmentation data-set and trained on the Detectron2 model. The main aim of this model is to perform instance segmentation to extract test strips from test cassettes. Figure \ref{fig:figure3}, showcase this aim. To build a robust model, we curated images using different mobile cameras, in different backgrounds, varying lighting conditions as well as random objects in the background. After taking raw images, we transformed them into a set resolution, 2582 x 1937, this is the most common resolution of a modern 5MP camera where images are in 4:3 aspect ratio. 

After gathering the data, we use the Labelme\cite{Wada_Labelme_Image_Polygonal} tool for annotation, as it supports both VOC and COCO formats. The annotation built using this tool on the images will be used to perform the instance segmentation. We transform our annotations into COCO format which are easily trainable.

\begin{figure}[!htb]
\centering
\includegraphics[width=0.6\textwidth]{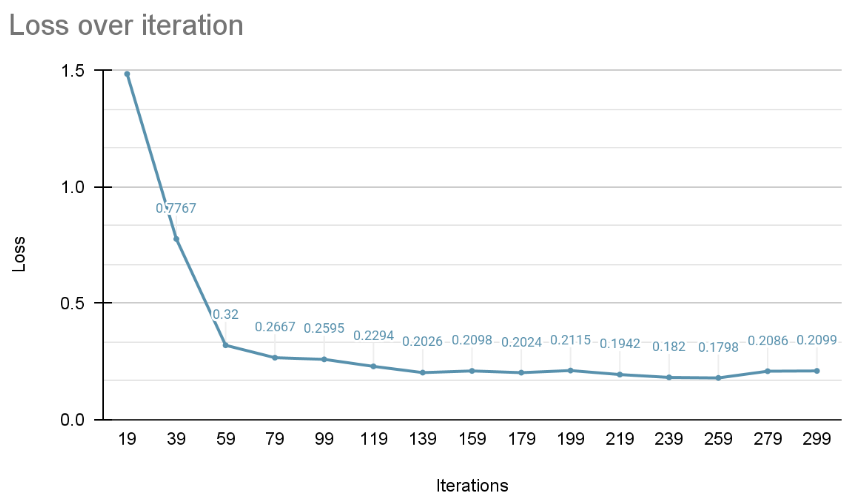}
\caption{\label{fig:figure4} Detectron2 COCO Instance Segmentation with Mask R-CNN model performance over Lateral flow test cassettes.}
\end{figure}
To build this model we are using COCO Instance Segmentation with Mask R-CNN model\cite{matterport_maskrcnn_2017}, this is done to build a robust model by using a pre-trained model which scores very highly,  41.0 on box Average Precision and 37.2 on mask Average Precision and thus reduce the dependence on training a very large dataset and while maintaining the performance of the model. Figure \ref{fig:figure4}, demonstrates the learning process of the model, as the loss decreases the model accuracy to determine the presence of test strips from the test cassettes increases. 
The model returns an impressive benchmark inference speed score of processing an image at an average of 0.19 with 5.18 fps.

\subsection{Binary and Multi Class Classification Model}

For these models, the required input image dataset is the test strips which are obtained from the previous Instance segmentation model. It becomes critical that the right training data is fed into each model. The effect of training data on loss function guides us through this. Binary Classifiers will be trained with all images partitioned into two classes, positive and unknown(negative and inconclusive). 

Multi-Class Classifiers will be trained with all the images being segregated into different classes such as Positive IGG, Positive IGM, Positive IGG and IGM, Negative or inconclusive. The defined architecture has 5 output classes with one neuron which provides the probability of each class. 

\begin{figure}[!htb]
\centering
\includegraphics[width=0.9\textwidth]{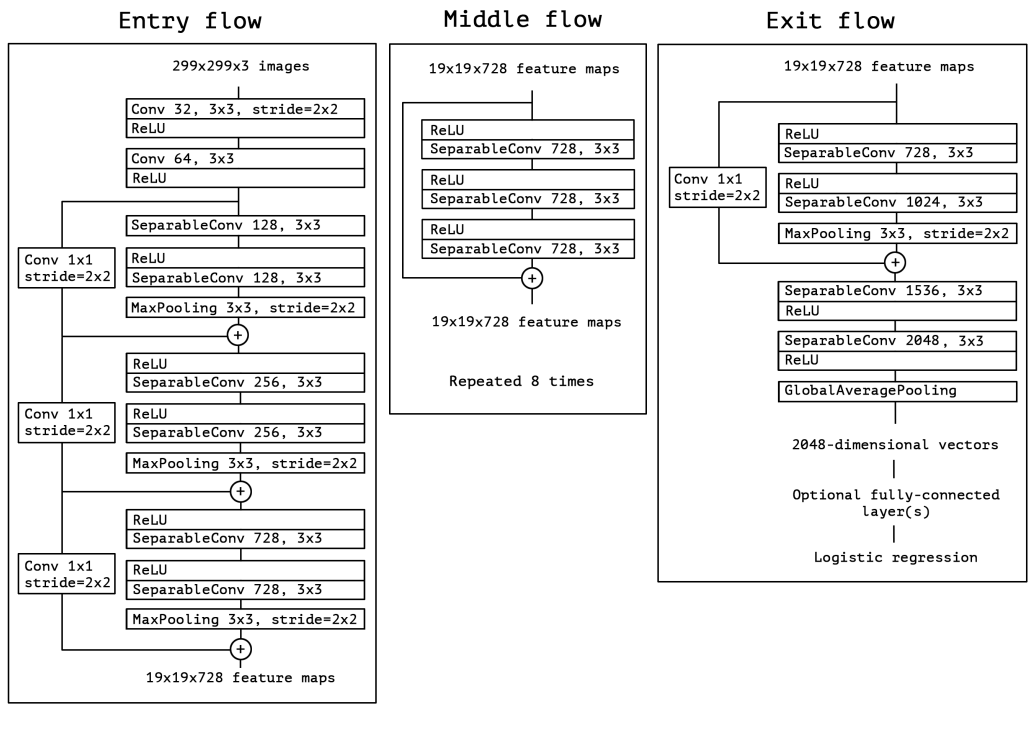}
\caption{\label{fig:figure5} Xception Model Architecture, the input embedding layer and end neuron layers are modified according to our data and expected result. The data first goes through the entry flow, then through the middle flow which is repeated eight times, and finally through the exit flow. Note that all Convolution and Separable Convolution layers are followed by batch normalization (not included in the diagram). All Separable Convolution layers use a depth multiplier of 1 (no depth expansion).}
\end{figure}

Each image is of 300 x 875 resolution and there are a total of 337 training, 85 validation and 17 test images for binary and multi-class classification models. Since we have a small size of data, we are using the Xception\cite{chollet2017xception} pre-trained model’s weights and fine-tune\cite{chollet2015keras} the existing model with our new data. Below figure \ref{fig:figure5} describe the configuration of the model.

\newpage
\begin{figure}[!htb]
\centering
\includegraphics[width=0.6\textwidth]{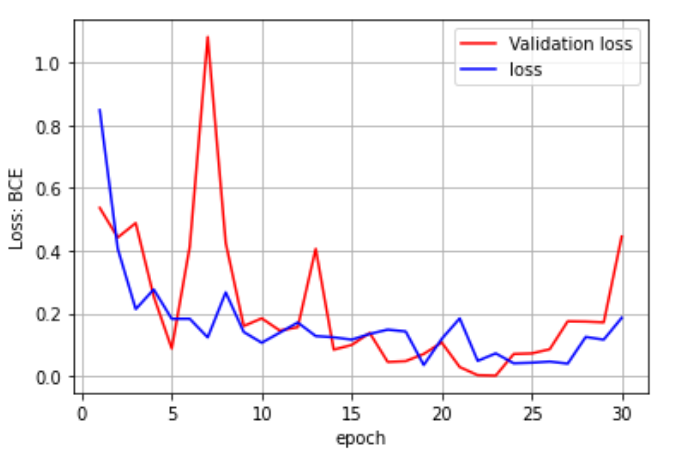}
\caption{\label{fig:figure6} Binary Classification model Loss over 30 epochs}
\end{figure}

In figure \ref{fig:figure6}, the binary cross-entropy loss of the model can be seen to have subtle variations on the validation set. After performing hyperparameter optimisation, in each epoch, loss measure can be seen decreasing with some exceptions. This implies that the model is learning but sometimes it has a tough time to generalise on unseen data sets while predicting the classes. The best weights are found at the 19th epoch and shown in Table 2.
\begin{table}[!htb]
\centering
\begin{tabular}{l|r|r|r}
\hline 
& \multicolumn{1}{|p{3cm}|}{\centering Train set \\ evaluation scores}   
& \multicolumn{1}{|p{3cm}|}{\centering Validation set \\ evaluation scores}
& \multicolumn{1}{|p{3cm}}{\centering Test set evaluation \\ evaluation scores}  \\\hline
Binary cross entropy & 0.202241 & 0.240638 & 0.194755 \\
Accuracy & 0.923630 & 0.912064 & 0.926810 \\
F1 score mean & 0.921999 & 0.912423 & 0.921435 \\
Mean Precision & 0.949316 & 0.937087 & 0.955327 \\
Mean Recall & 0.905966 & 0.898664 & 0.902135 \\
\hline
\end{tabular}
\caption{\label{tab:widgets}Binary Classification model performance measures}
\end{table}

The binary classification model has good performance on train, validation and test datasets. The values of loss and metrics can be seen to be similar in these datasets. This tells that the model is not overfitting on the dataset. The f1 score of 0.921 on the validation dataset is acceptable.
\begin{figure}[!htb]
\centering
\includegraphics[width=0.6\textwidth]{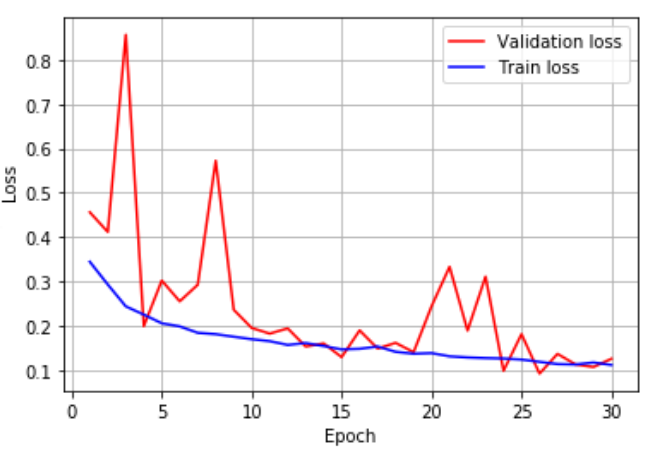}
\caption{\label{fig:figure7} Multi-Class Classification Loss over 30 epochs}
\end{figure}

In figure \ref{fig:figure7}, the categorical cross-entropy loss of the model is initially struggling to converge over the validation dataset for the multi-class classifier as we see a large variation in the 0-10 epoch. But it later improves and generalisation is overall good.
\begin{table}[!htb]
\centering
\begin{tabular}{l|r|r|r}
\hline 
& \multicolumn{1}{|p{3cm}|}{\centering Train set \\ evaluation scores}   
& \multicolumn{1}{|p{3cm}|}{\centering Validation set \\ evaluation scores}
& \multicolumn{1}{|p{3cm}}{\centering Test set evaluation \\ evaluation scores}  \\\hline
Categorical cross entropy & 0.081054 & 0.092119 & 0.094178 \\
Accuracy & 0.968118 & 0.962500 & 0.965517 \\
F1 score mean & 0.940510 & 0.929417 & 0.936398 \\
Mean Precision & 0.945815 & 0.929264 & 0.941134 \\
Mean Recall & 0.937232 & 0.931588 & 0.93854 \\
\hline
\end{tabular}
\caption{\label{tab:tab3}Multi-Class Classification model performance measures}
\end{table}
The best weights were found at the 18th epoch and Table \ref{tab:tab3}, shows the performance of the model. In conclusion, the multi-class classification model is generalizing well on the unseen data, as the values of the evaluation test set and validation sets are closer to the train set.

\subsection{Segmentation Model}

The Segmentation models are used to find the presence of a certain type of class. To prepare the dataset for this model, we use the VGG Image Annotator (VIA)\cite{dutta2019vgg} tool and create masks for all the images. We use the pre-trained model, for Legendary UNet\cite{ronneberger2015unet} with the EfficientNetB1\cite{tan2020efficientnet} backbone model’s weight. For this model we are using the Dice Coefficient\cite{10.2307/1932409} as the performance metric, to gauge similarity of two samples. The Dice coefficient can be used to compare the pixel-wise agreement between a predicted segmentation and its corresponding ground truth. The formula is given by:
\[C = \frac{2 * | X {\displaystyle \cap } Y | }{|X| + |Y|}\]
where X is the predicted set of pixels and Y is the ground truth. The Dice coefficient is defined to be 1 when both X and Y are empty. Our objective is to maximize the Dice coefficient and identify/locate the type of defect present in the image.
\begin{figure}[!htb]
\centering
\includegraphics[width=0.6\textwidth]{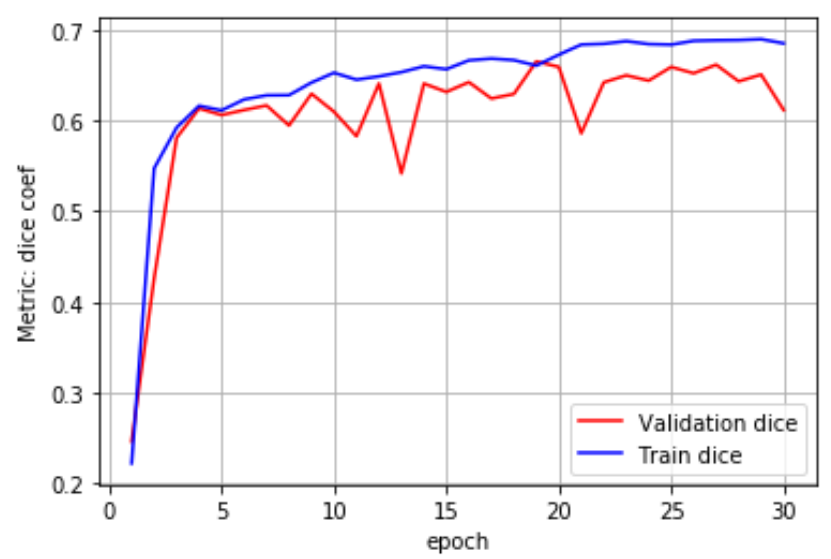}
\caption{\label{fig:figure8}  Positive-IGG class Segmentation model dice coefficient over epochs}
\end{figure}
In figure \ref{fig:figure8}, the dice coefficient stabilizes around 65\% after 15 epochs for class 1 which is to detect Positive IGG. This score is good in terms of the number of images for this class.

\begin{table}[!htb]
\centering
\begin{tabular}{l|r|r|r}
\hline 
& \multicolumn{1}{|p{3cm}|}{\centering Train set \\ evaluation scores}   
& \multicolumn{1}{|p{3cm}|}{\centering Validation set \\ evaluation scores}
& \multicolumn{1}{|p{3cm}}{\centering Test set evaluation \\ evaluation scores}  \\\hline
Dice loss & 0.285742 & 0.334879 & 0.388797 \\
Dice coefficient & 0.714258 & 0.665121 & 0.611203 \\
\hline
\end{tabular}
\caption{\label{tab:tab4}Positive-IGG class Dice performance scores}
\end{table}

The Dice coefficient can be considered as an F1 score to evaluate the performance metrics of the segmentation models. In Table \ref{tab:tab4}, for class Positive-IGG, an acceptable score of 0.665 is obtained on the validation data set.

\begin{figure}[!htb]
\centering
\includegraphics[width=0.6\textwidth]{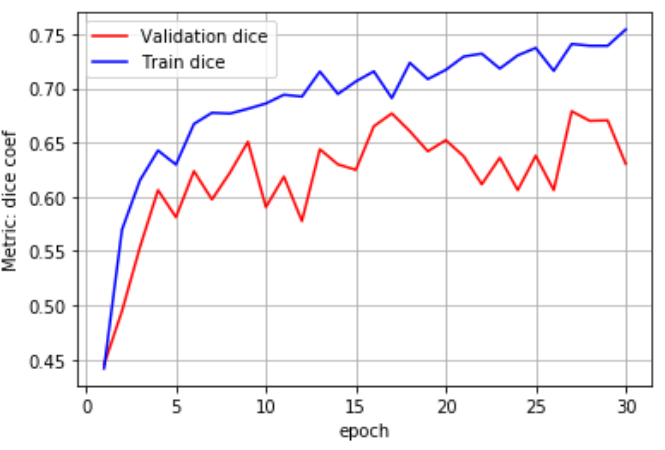}
\caption{\label{fig:figure9}  Positive-IGM class Segmentation model dice coefficient over epochs.}
\end{figure}

In figure \ref{fig:figure9}, the dice coefficient stabilizes around 65\% at 20 epoch for class 2 which is to detect Positive IGM. This score is good in terms of the number of images for this class.

\begin{table}[!htb]
\centering
\begin{tabular}{l|r|r|r}
\hline 
& \multicolumn{1}{|p{3cm}|}{\centering Train set \\ evaluation scores}   
& \multicolumn{1}{|p{3cm}|}{\centering Validation set \\ evaluation scores}
& \multicolumn{1}{|p{3cm}}{\centering Test set evaluation \\ evaluation scores}  \\\hline
Dice loss & 0.295981 & 0.308811 & 0.338572 \\
Dice coefficient & 0.766948 & 0.678812 & 0.655394\\
\hline
\end{tabular}
\caption{\label{tab:tab5}Positive-IGM class Dice performance scores}
\end{table}

The Dice coefficient can be considered as an F1 score to evaluate the performance metrics of the segmentation models. In Table \ref{tab:tab5}, for class Positive-IGG, an acceptable score of 0.665 is obtained on the validation data set.

Similarly, we calculate the Dice coefficient for Positive-IGG and IGM class and Negative class and inconclusive class. Below table \ref{tab:tab6}. summarises the Dice coefficient for all the classes. 

\begin{table}[!htb]
\centering
\begin{tabular}{l|r|r|r|r|r}
\hline 
Dataset
& \multicolumn{1}{|p{2cm}|}{\centering Positive IGG \\ class}   
& \multicolumn{1}{|p{2.1cm}|}{\centering Positive IGM \\ class}
& \multicolumn{1}{|p{2cm}|}{\centering Positive IGG \\ \& IGM class}    
& \multicolumn{1}{|p{2cm}|}{\centering Inconclusive  \\ class}
& \multicolumn{1}{|p{2cm}}{\centering Negative  \\ class}  \\\hline
X Train & 0.714258 & 0.766948 & 0.735519 & 0.67421 & 0.821943\\
X Validation & 0.665121 & 0.678812 & 0.709641 & 0.63597 & 0.76066\\
X Test & 0.611203 &0.655394 & 0.698548 & 0.619753 & 0.78822\\
\hline
\end{tabular}
\caption{\label{tab:tab6}Summary of Dice Coefficient on Unet - EfficientNetB1 model for different classes on data splits.}
\end{table}
\subsection{Active Learning}

This is an addition to the existing classification models. The role of active learning\cite{roy2001toward} is to improve the overall performance of binary and multiclass classifiers. Using the Active Learning framework we allow increasing classification performance by intelligently querying the label for the most informative instances.

\begin{figure}[!htb]
\centering
\includegraphics[width=0.8\textwidth]{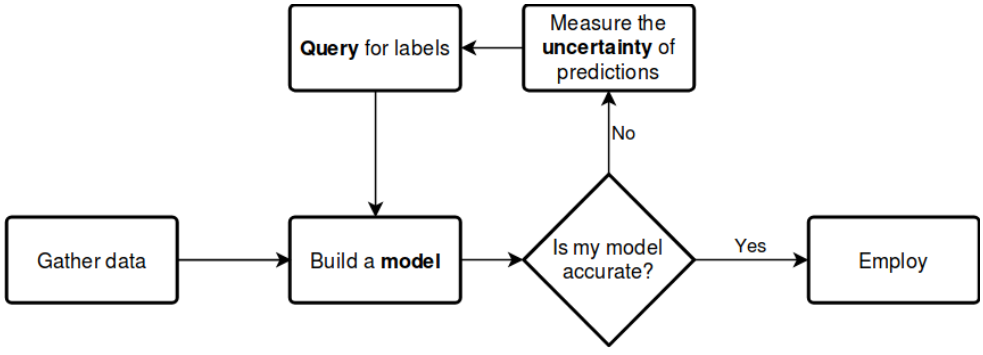}
\caption{\label{fig:figure10}  Active Learning on Binary and Multi-class classifiers. Souce: modal}
\end{figure}

We are using the modAL framework\cite{modAL2018} to implement Active Learning, the working is described in figure \ref{fig:figure10}. The key components are choosing the uncertain measure and query strategy for the labels. We are handling the first component by asserting the probability score of each of the classes of binary and multi-class classifiers at a set threshold. This threshold is derived through rigorous testing of the overall system. The second component is handled using a human-in-the-loop response or a pre-built framework of modAL.

\section{Inference and Results}

\begin{figure}[!htb]
\centering
\includegraphics[width=1\textwidth]{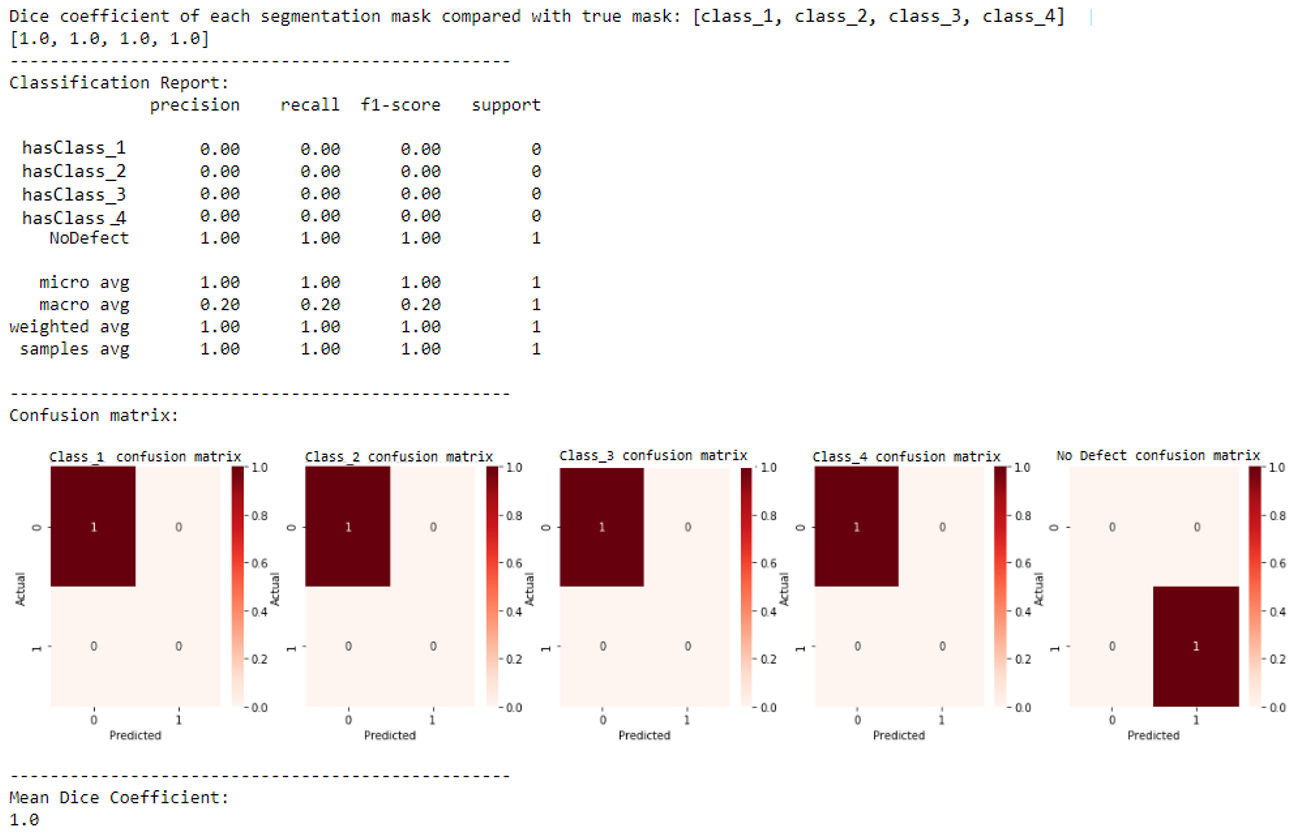}
\caption{\label{fig:figure11}  Negative class evaluation on Multi-class classifier and Segmentation Models}
\end{figure}

From the training above, the best models are saved to make inferences on images. We define the different dependencies for loading saved models as it is important while we use custom metrics. We define a custom method for generating predictions using Classifier models and another custom method for generating predictions using Segmentation models. Different thresholds such as Area thresholds and Classification thresholds are applied to the predictions of the models. Figure \ref{fig:figure11}, shows sample evaluation on a single image. The image has a negative class(no defect) and the models have perfectly detected that image has a negative class(no defect).

\begin{table}[!htb]
\centering
\begin{tabular}{l|r|r|r|r|r}
\hline 
Dataset
& \multicolumn{1}{|p{2cm}|}{\centering Class 1 \\ Positive IGG}   
& \multicolumn{1}{|p{2.1cm}|}{\centering Class 2\\ Positive IGM}
& \multicolumn{1}{|p{2cm}|}{\centering Class 3 \\Positive IGG \& IGM}    
& \multicolumn{1}{|p{2cm}|}{\centering Class 4 \\ Inconclusive}
& \multicolumn{1}{|p{2cm}}{\centering Class 5 \\ Negative}  \\\hline
X Train & 0.9524 & 0.9823 & 0.8203 & 0.8724 & 0.9787\\
X Validation & 0.9479 & 0.9806 & 0.8162 & 0.8611 & 0.9738\\
X Test & 0.9442 & 0.9826 & 0.8047 & 0.8341 & 0.9787\\
\hline
\end{tabular}
\caption{\label{tab:tab7}Mean Dice Coefficient class wise.}
\end{table}

From the table \ref{tab:tab7}, we can understand that each metric can be used to see the fact that the model did not overfit on the train set, validated well on validation dataset and generalising well in unseen test set. 
\begin{figure}[!htb]
\centering
\includegraphics[width=1\textwidth]{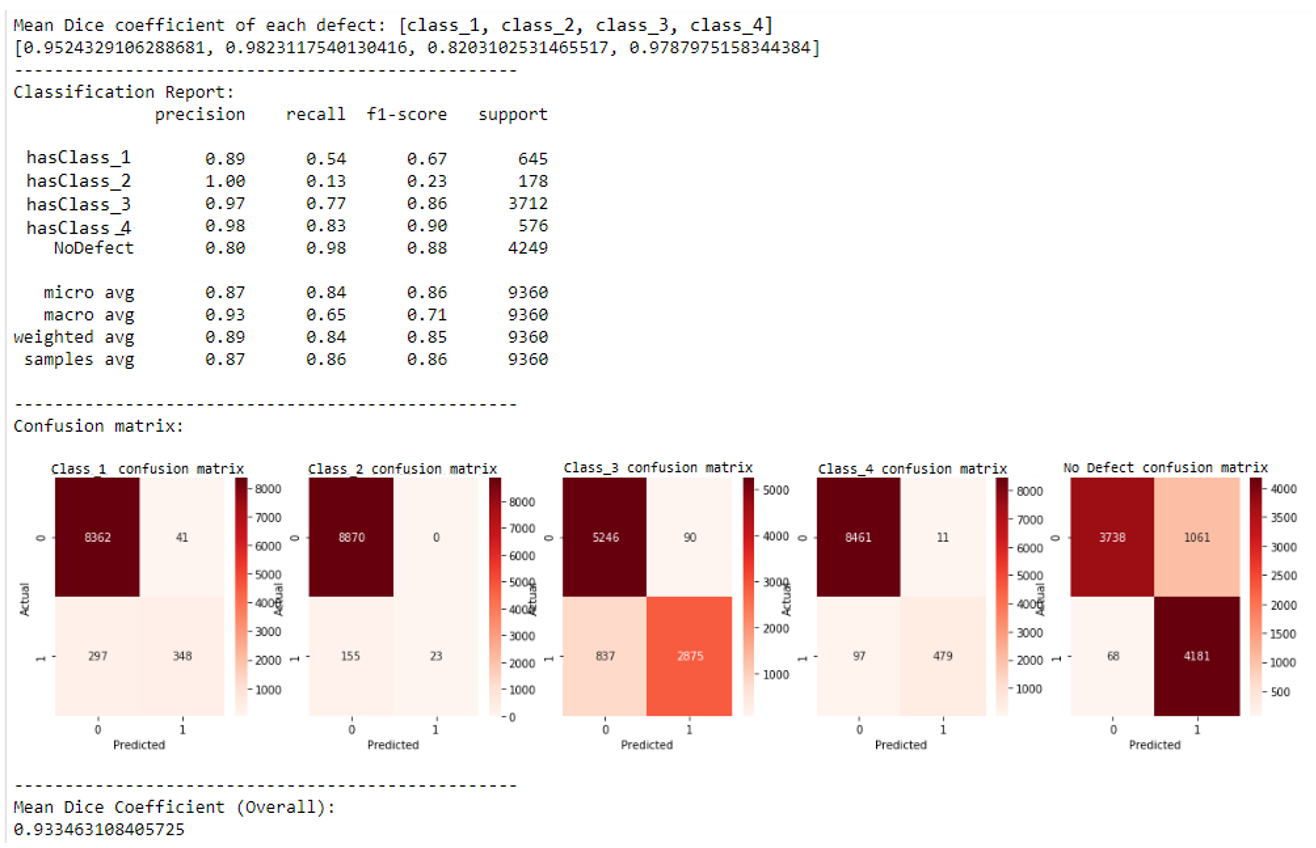}
\caption{\label{fig:figure12}  Overall evaluation scores on all the Training images on Segmentation Model, Note: Some images are added by transformations and noise insertion techniques.}
\end{figure}

In figure \ref{fig:figure12}, we observe the classification report for the multi-class classifier on augmented images. Due to data augmentation, the F1 scores are low but overall precision and recall are good. The mean Dice coefficient for overall classes is perfect and is around 93\% and the mean Dice coefficient for each class is above 85\%. Thus, if the Multi-class classifier produces an inaccurate result, these can be well covered up by using segmentation models. 

\section{Conclusion}

In this study we have carried out an analysis of accuracy on machine learning model to evaluate the results of lateral flow devices tests. The image reader demonstrate a novel AL algorithm can perform and scale well across different vendors with 84\% sensitivity and 87\% specificity. The model built, is self-adaptive to learn new images with human-in-the-loop active learning model present and the current values of sensitivity and specificity increases with each new inference when validated by human. This overall improves the false positive rates and maintains true positive rate as the results are identical with each iteration with active learning. Another advantage is the processing time is below 2 seconds and it remains constant with increase in load, thus it scales vertically and horizontally. Additionally, due to the presence of segmentation model, the model excellently detects faint color bands which are due to low vial load and this likely help users who are unclear about reading LFDs and are prone to make mistake likely or otherwise. One disadvantage is the quality of image, we have considered the current trend of available smartphone collect images above 5 mega pixels if the images are below this metric or have issues such as blur, brightness, reflection or darkness, the model would likely reject such images. Apart from this limitation, the model works at a level of consistent performance with the capacity of processing millions of images rapidly with feature to improve the ineffectualness with active learning, thus evolving better every time, and reducing the user error and increasing overall accuracy of model.


\bibliographystyle{alpha}
\bibliography{sample}

\end{document}